\documentclass[a4paper,fleqn]{cas-dc}
\usepackage{color}
\usepackage{makecell}
\usepackage[numbers]{natbib}
\usepackage[justification=centering]{caption}
\def\tsc#1{\csdef{#1}{\textsc{\lowercase{#1}}\xspace}}
\tsc{WGM}
\tsc{QE}
\tsc{EP}
\tsc{PMS}
\tsc{BEC}
\tsc{DE}

\begin{document}
\begin{sloppypar}
\let\WriteBookmarks\relax
\def\floatpagepagefraction{1}
\def\textpagefraction{.001}
\shorttitle{Multi-Task Learning for Fatigue Detection and Face Recognition of Drivers}
\shortauthors{Shulei Qu et~al.}

\title [mode = title]{Multi-Task Learning for Fatigue Detection and Face Recognition of Drivers via Tree-Style Space-Channel Attention Fusion Network}                      

\tnotetext[1]{This document is the results of the research project funded by the National Science Foundation of China under Grant 62372190 and Industry University Cooperation Project of Fujian Province under Grant 2021H6030.}


\author[1,2]{Shulei Qu}

\credit{Conceptualization of this study, Methodology, Software}

\author[1,2]{Zhenguo Gao}[orcid=0000-0003-3115-6959]
\ead{gzg@hqu.edu.cn}
\cormark[1]
\credit{Supervision, Writing and editing}


\author[1,2]{Xiaowei Chen}
\credit{Writing and reviewing}

\author[3,2]{Na Li}
\credit{Writing and reviewing}

\author[1,2]{Yakai Wang}
\credit{Writing and reviewing}

\author[1,2]{Xiaoxiao Wu}
\credit{Writing and reviewing}


\cortext[cor1]{Corresponding author}
\address[1]{Department of Computer Science and Technology, Huaqiao University, Xiamen 361021, Fujian, China}
\address[2]{Key Laboratory of Computer Vision and Machine Learning of Fujian  Provincial Universities, Xiamen 361021, Fujian, China}
\address[3]{Department of Mechanical Engineering and Automation, Huaqiao University, Xiamen 361021, Fujian, China}


\begin{abstract}
In driving scenarios, automobile active safety systems are increasingly incorporating deep learning technology. These systems typically need to handle multiple tasks simultaneously, such as detecting fatigue driving and recognizing the driver's identity. However, the traditional parallel-style approach of combining multiple single-task models tends to waste resources when dealing with similar tasks. Therefore, we propose a novel tree-style multi-task modeling approach for multi-task learning, which rooted at a shared backbone, more dedicated separate module branches are appended as the model pipeline goes deeper. Following the tree-style approach, we propose a multi-task learning model for simultaneously performing driver fatigue detection and face recognition for identifying a driver. This model shares a common feature extraction backbone module, with further separated feature extraction and classification module branches. The dedicated branches exploit and combine spatial and channel attention mechanisms to generate space-channel fused-attention enhanced features, leading to improved detection performance. As only single-task datasets are available, we introduce techniques including alternating updation and gradient accumulation for training our multi-task model using only the single-task datasets. The effectiveness of our tree-style multi-task learning model is verified through extensive validations.
\end{abstract}




\maketitle

\section{Introduction}

In the driving scenario, as automobile active safety systems increasingly embrace deep learning technology, more deep learning techniques are being applied in automobiles. Mature deep learning technologies are typically designed for targeted tasks, necessitating the deployment of numerous models within an automotive system to accomplish various tasks. Previous research has focused on optimizing the performance of individual tasks and has proposed many deep learning-based methods. 

When facing multiple tasks, as the approach of parallelly combining multiple single-task models is simple and straight forward, so this approach is usually adopted in the initial of multi-task learning paradigm, which is referred to as parallel-style approach. However, tasks in driving scenarios often exhibit some strong similarities. For example, commercial vehicles such as buses need to monitor both the driver's fatigue status and their identity, both of which require extracting features from the driver's facial images.

Although excellent algorithms exist for both fatigue detection and face recognition, the parallel-style approach for the joint task of fatigue detection and face recognition would lead to significant redundant computations and resource wastage due to the similarity between these tasks. As both tasks rely on the facial images of the driver for assessment, there must be shared underlying features between these two tasks. Therefore, more efficient multi-task learning model is required.

Furthermore, most datasets in the literature are initially constructed and annotated for a targeted task, whereas multi-task learning requires datasets with multiple annotations for various tasks, making them usually unable to be trained using single-task datasets. 

In this paper, we deal with the two closely related tasks of fatigue driving detection and face recognition of drivers. Considering that both fatigue driving detection and face recognition require the same input image, we propose a novel \textbf{tree-style approach} for multi-task modeling. In this modeling approach, all tasks share a common feature extraction network backbone as the root module, and as the model pipeline goes deeper, dedicated module branches are appended, with the following branches being more dedicated to the final tasks. Finally, for each task, a dedicated task head is appended. Thus, the model pipeline forms a tree structure. The entire tree is referred to as a tree-style multi-task model, and the model part corresponding to a node in the tree is called a module. The model is sometimes called a tree-model for emphasizing its tree-structure. Currently, there isn't a tree-style multi-task model targeted for fatigue detection and face recognition of drivers.    

In addition, existing datasets are initially constructed and labeled for one dedicated task. To train our multi-task model using existing single-task datasets, we adopted two training techniques including alternately updating and gradient accumulation. Alternately updating was used to update the weights of the generator and the discriminator in GAN models\cite{gan}. Gradient accumulation was initially used to address out-of-memory issues by imitating the training process with large batches through the accumulation of gradients over multiple small batches, thus reducing memory consumption.

Following a tree-style approach, we propose a multi-task learning model to simultaneously perform driver fatigue detection and face recognition for driver identification. A Convolutional Neural Network (CNN) based backbone serves as the root module, shared by both tasks. Then, following the root module are two branch modules, which are dedicated to fatigue driving detection and face recognition of drivers, respectively. The two branch modules have similar structures but with different parameters. Each branch module contains a LANet, a SENet, and a residual connection as three parallel branches, thus it is named LASE-Net. In a LASE-Net module, the LANet module is dedicated to emphasizing channel attention, while the SENet module tries to exploit spatial attention. Thus, both channel attention and spatial attention are employed to obtain better features. The residual connection facilitates the training process for better convergence. Our entire tree-style model is named a Tree-style Space-Channel Attention Fusion (T-SCAF) model. Extensive experiments were conducted, and the results show that our T-SCAF model achieves state-of-the-art performance on average.

The contributions in this paper are summarized as follows:
\begin{itemize}
\item We propose a tree-style multi-task modeling approach for multi-task learning, where all tasks share a common feature extraction network backbone as the root module, and as the model pipeline goes deeper, more dedicated separate modules are branched further, with the following branches being more dedicated to the tasks.

\item We propose a Tree-style Space-Channel Attention Fusion (T-SCAF) model for simultaneously conducting the two tasks of driver fatigue detection and driver identification.

\item We introduce alternating updation and gradient accumulation techniques to train our T-SCAF model, enabling it to be trained using single-task datasets.

\end{itemize}

The remaining sections are organized as follows. Section~\ref{sec_rel_work} reviews related work. Section~\ref{sec_method} provides our T-SCAF model; Section~\ref{sec_evaluation} introduces the alternating updation and gradient accumulation techniques for training T-SCAF. Section~\ref{sec_evaluation} provides the experiment results. Finally, Section~\ref{sec_conclusion} presents the conclusion. 

\section{Related Work}
\label{sec_rel_work}
\subsection{Fatigue Detection}
Fatigue detection methods based on driver behavior characteristics aim to discern whether a driver is experiencing fatigue by analyzing the driver's facial images. This is primarily achieved by identifying features such as eye movements, mouth movements, head posture, and facial expressions. Several studies have proposed deep learning models specifically for fatigue detection, achieving promising performance.

Ref.~\cite{real-time} introduced a dual-stream network, where each stream extracts two feature steams from respectively of the left eye and mouth images parts. They fused the features extracted from these four streams to generate detection results, achieving an accuracy of 91.3\% on their self-collected dataset. Ref.~\cite{husain2022development} utilized ResNext101 as the backbone network and added a Weibull Pooling layer afterward. This model extracts features from input facial images and performs classification, achieving an accuracy of 84.21\%.

\subsection{Face Recognition}
Face recognition is a biometric authentication technology aimed at confirming an individual's identity by analyzing and identifying biological features in the facial image. Face recognition models extract features from facial images by employing convolutional neural networks, then map these features into high-dimensional vectors. Typically, the similarity between different facial images is determined by calculating the distance (e.g., Euclidean distance, cosine distance) between their feature vectors, allowing for the identification of whether two facial images belong to the same person.

FaceNet\cite{facenet} introduced a novel task which adopts three images as input: two images of the same person from different angles, while the remaining image contains a different person. The objective is to minimize the feature distance between images of the same person and maximize the distance between features of different individuals. ArcFace\cite{arcface} improves upon softmax loss by introducing the ArcFace loss, which expands the inter-class margin during classification training, which enhances recognition accuracy.

\subsection{Multi-Task Learning}
Multi-task learning models typically consist of multiple branches, with each dedicated to a specific task. The former (shallower) layers of a model are expected to extract shared feature representations, while a task-dedicated branch is responsible for outputting predictions for its targeted task. This allows the shared representations at the former layers to be further-exploited at later (deeper) layers through the optimization process of targeted tasks, enhancing the model's ability to adapt to each task. Based on how the model parameters are shared, multi-task models are categorized into soft parameter sharing type and hard parameter sharing type.

In soft parameter sharing type multi-task model, each task typically has its independent module. However, the modules of different tasks share parameters of certain layers through various mechanisms, as demonstrated in works\cite{SNR,ma2018modeling}. This type of multi-task model corresponds to what we refer to as the parallel-style modeling approach.

In hard parameter sharing type multi-task model, the former layers of the model (typically the first few layers) are shared, while each task has its independent later-layer branch module. This means that each task shares the same feature extractor, but there are separate branch modules containing task-specific layers for generating task-specific outputs, as seen in works\cite{Multi-task_Learning_Using_Uncertainty,gradnorm}. This type of multi-task model corresponds to what we refer to as the tree-style modeling approach, yet their tree structures are restricted to 2-layer depth.

Models such as Mask R-CNN\cite{mask_rcnn} simultaneously perform tasks like image segmentation, object recognition, and classification. Similarly, MTCNN\cite{Joint_Face_Detection} and RetinaFace\cite{retinaface} can detect faces and facial landmarks concurrently.

\section{The T-SCAF Model for Fatigue Detection and Face Recognition of Drivers}
\label{sec_method}
\subsection{Tree-style Multi-Task Modeling Approach}
We propose a conceptual tree-style multi-task modeling approach to multi-task learning. In this approach, all tasks share a common feature extraction network backbone as the root module. As the model pipeline progresses, more dedicated separate modules branch out further, with subsequent branches being more specialized for the final tasks. Finally, each task has a dedicated task head. Thus, the model pipeline forms a tree.

In the tree-style multi-task modeling approach, the branching points depend on the similarity levels between related tasks as well as the model structures for the tasks. The tasks may be grouped hierarchically, which can be directly mapped to the tree structure of a tree-style multi-task model.

Our tree-style multi-task modeling approach contains the so-called hard parameter sharing type multi-task learning as a special case, where the depth of the tree-style model equals two. 

\subsection{Tree-style Space-Channel Attention Fusion (T-SCAF) model}
\begin{figure*}
    \centering
    \includegraphics[width=1\linewidth]{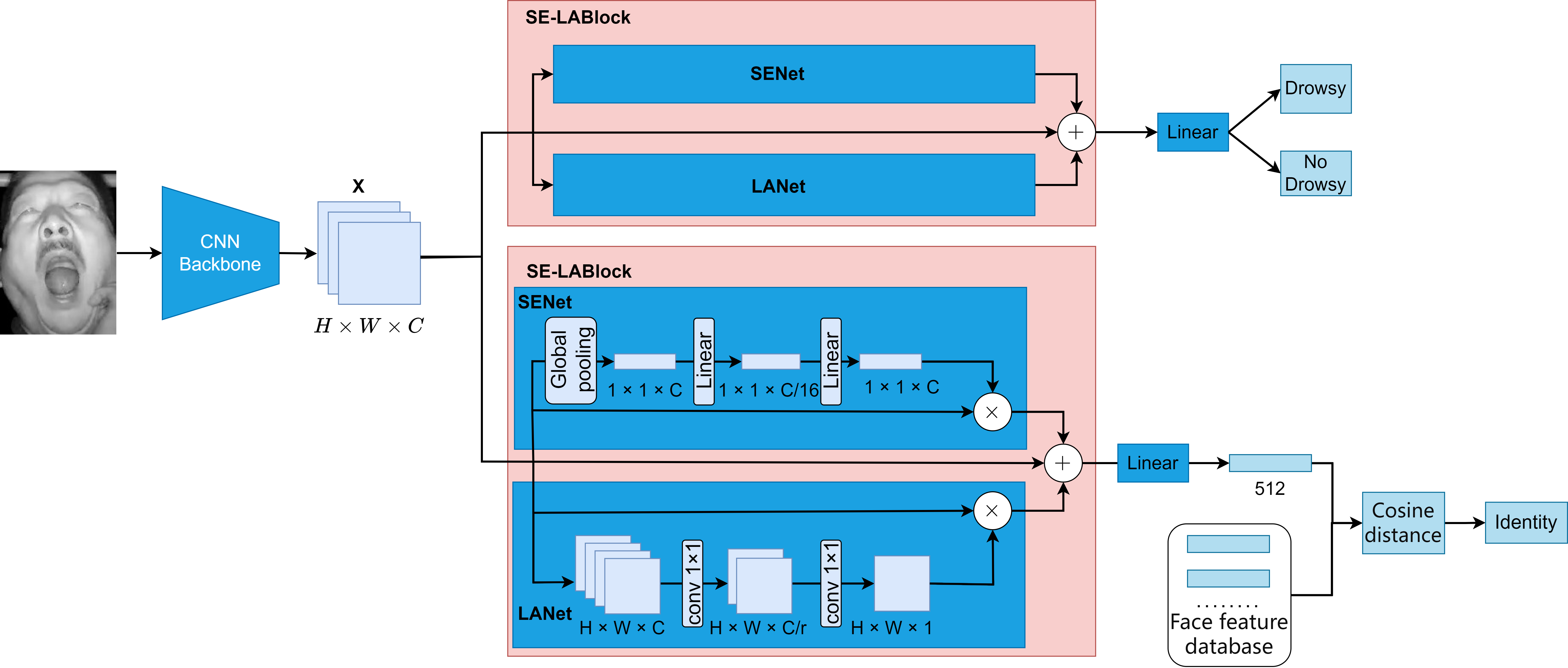}
    \caption{Structure of our tree-style space-channel attention fusion (T-SCAF) model for fatigue detection and face recognition of drivers.}
    \label{fig:structure}
\end{figure*}

Focusing on the two closely related tasks of fatigue detection and face recognition of drivers, adopting the tree-style multi-task modeling approach, we propose our Tree-style Space-Channel Attention Fusion (T-SCAF) model, which exploits and fuses both space and channel attention mechanisms to produce more representative features for promoting the detection performance. The structure of the T-SCAF model is illustrated in Fig.~\ref{fig:structure}.

We use the same backbone network as the root module of the tree-model, which is used to extract feature matrices from images for both fatigue detection and face recognition tasks. Thus, both tasks share the same feature matrix generated by the backbone module. The root module is followed by two branch modules, each of which contains a LANet module, a SENet module, and a residual connection, and hence such a branch module is called a LASE-Net module. The two LASE-Net branch modules are respectively dedicated to fatigue detection and face recognition. In a LASE-Net module, the LANet module is dedicated to emphasizing space attention, while the SENet module tries to exploit channel attention, thus both space attention and channel attention are employed to obtain more representative features. The residual connection facilitates the training process for better convergence. 

\subsection{The LANet Module}
The LANet module is proposed in \cite{ls_cnn}, here it is adopted to achieve space attention. LANet is a convolutional structure composed of 2 layers of $1{\times}1$ convolutions followed by ReLU and Sigmoid activation functions,  respectively. The pipeline of the SENet is illustrated in the lower part of of the lower LASE-Net box in Figure~\ref{fig:structure}. 

The input feature matrix $\textbf{x}_1{\in}\mathbb{R}^{H{\times}W{\times}C}$ first passes through the a convolutional layer, reducing the channel dimension of the feature matrix to $C/r$, where $r$ is a hyper-parameter defining the feature channel compression ratio. Then the reduced features are passed to ReLU activation functions. Next, the features pass through the second convolutional layer to resize to $H{\times}W{\times}1$, thus reducing the number of feature channels to 1. The feature matrix passes through the Sigmoid layer to obtain the spatial attention weight matrix $\textbf{Att}_\text{space}{\in}\mathbb{R}^{H{\times}W{\times}1}$. Till now, all the channels of a pixel in the input feature matrix is compressed into a scale value in the final matrix, thus the scale value reveals the relative importance of each pixel in some sense. As a result, the final $\textbf{Att}_\text{space}{\in}\mathbb{R}^{H{\times}W{\times}1}$ encloses the space attention information.   

The obtained spatial attention weight matrix $\textbf{Att}_\text{space}$ is duplicated along the channel dimension to change shape to $H{\times}W{\times}C$, which is also the shape of $\textbf{x}_1$. Weighting the elements in $\textbf{x}_1$ using $\textbf{Att}_\text{space}$, i.e., by element-wise multiplication of $\textbf{x}_1$ and $\textbf{Att}_\text{space}$, we obtain the space-attention adjusted feature matrix $\textbf{x}_\text{LANet}$. 

In summary, the LANet module generates $\textbf{x}_\text{LANet}$ from $\textbf{x}_1$ following Eq.~\eqref{eq_lanet_e2e}.

\begin{subequations}
\label{eq_lanet_e2e}
\begin{equation}
\label{eq_lanet_e2e_1}
\textbf{x}_\text{LANet}{=}\text{dup}\left(\textbf{Att}_\text{space},[3],[C]\right){\odot}\textbf{x}_1.
\end{equation}
\begin{equation}
\label{eq_lanet_e2e_2}
\textbf{Att}_\text{space}{=}\sigma\left(\text{CNN}\left(\text{ReLU}\left(\text{CNN}\left(\textbf{x}_1\right)\right)\right)\right).
\end{equation}
\end{subequations}

In Eq.~\eqref{eq_lanet_e2e}, $\text{dup}(\cdot,[i_1,i_2,\ldots,i_{n}],[j_1,j_2,\ldots,j_{n}])$ represents the replication function to enlarge a matrix by enlarging the $i_k$-th dimention by $j_k$ times, $k{\in}\{1,2,\ldots,n\}$. Here in Eq.~\eqref{eq_lanet_e2e}, the matrix is enlarged from size $H{\times}W{\times}1$ to $H{\times}W{\times}C$. $\sigma(\cdot)$ and $\mathrm{ReLU}(\cdot)$ represent the corresponding activation functions, $\text{CNN}(\cdot)$ denotes the function of a convolution layer, $\mathrm{avg}(\cdot)$ represents the global average pooling function, and $\odot$ represents the Hadamard product.

\subsection{The SENet Module}
The SENet module is proposed in \cite{sen}, here it is adopted to achieve channel attention. SENet is a fully connected network consisting of one global average pooling layer, two fully connected layers, and ReLU and Sigmoid activation function layers. The pipeline of the SENet is illustrated in the upper part of the lower LASE-Net box in Figure~\ref{fig:structure}. 

The input feature matrix $\textbf{x}_1{\in}\mathbb{R}^{H{\times}W{\times}C}$ first passes through the global average pooling layer, resulting in a new feature matrix with size $\textbf{x}_2{\in}\mathbb{R}^{1{\times}1{\times}C}$. The $m$-th channel of $\textbf{x}_2$ is calculated as Eq.~\eqref{eqen}.

\begin{equation}
\label{eqen}
\textbf{x}_{2,m}=\text{avg}(\textbf{x}_1){=}\frac{1}{H{\times}W}\left(\sum_{i=0}^H\sum_{j=0}^{W}x{i,j,m}\right)
\end{equation}

The feature matrix $x_2$ then passes through the first fully connected layer, resulting in a new feature matrix $x_3$ with a reduced channel dimension of $C/16$. $x_3$ then passes through the ReLU activation layer. Subsequently, the feature matrix passes through the second fully connected layer and the Sigmoid activation layer, resulting in a channel attention weight vector $\textbf{Att}_{channel}$ of size $1{\times}1{\times}C$. Till now, each feature slice in the feature matrix $x_1$ is compressed into a scale value in $\textbf{Att}_\text{channel}$, thus the scale value reveals the relative importance of each channel in some sense. As a result, $\textbf{Att}_\text{channel}{\in}\mathbb{R}^{1{\times}1{\times}C}$ encloses the channel attention information.    

$\textbf{x}_\text{channel}$ is duplicated to size of $H{\times}W{\times}C$ to obtain a channel attention matrix $\textbf{x}_4$. $\textbf{x}_4$ and $\textbf{x}_1$ have the same shape, so they are combined into a channel-attention adjusted feature matrix $\textbf{x}_\text{SENet}$ by element-wise multiplication. 
In summary, the SENet module generates $\textbf{x}_\text{SENet}$ from $x_1$ following Eq.~\eqref{eq_senet_e2e}. 

\begin{subequations}
\label{eq_senet_e2e}
\begin{equation}
\label{eq_senet_e2e_1}
\textbf{x}_\text{SENet}{=}\text{dup}\left(\textbf{Att}_\text{channel},[1,2],[H,W]\right){\odot}\textbf{x}_1
\end{equation}
\begin{equation}
\label{eq_senet_e2e_2}
\textbf{Att}_\text{channel}{=}\sigma\left(\text{FC}\left(\text{ReLU}\left(\text{FC}\left(\text{avg}(\textbf{x}_1)\right)\right)\right)\right).
\end{equation}
\end{subequations}

In Eq.~\eqref{eq_senet_e2e}, the replication function $\text{dup}_(\cdot)$ enlarges the shape size of the input matrix from $1{\times}1{\times}C$ to $H{\times}W{\times}C$. $\mathrm{FC}(\cdot)$ denotes the function of the full connection layer, $\mathrm{avg}(\cdot)$ represents the global average pooling function.

\subsection{Space-Channel Attention Fusion Module LASE-Net}

The original feather matrix $x_1$ not only passes the parallel LANet and SENet branches respectively to generate two attended feature matrix $\textbf{x}_\text{LANet}$ and $\textbf{x}_\text{SENet}$, but also directly connects to the end of the two branches via a residual connection. $\textbf{x}_\text{LANet}$, $\textbf{x}_\text{SENet}$, and $\textbf{x}_1$ are element-wisely added together to obtain a final space-channel attention fused feature matrix $\textbf{x}_\text{AttFused}$ with shape $H{\times}W{\times}C$, as expressed in Eq.~\eqref{eq_threebranch_comb}. 

\begin{equation}
\label{eq_threebranch_comb}
\textbf{x}_\text{AttFused}{=}\textbf{x}_1{+}\textbf{x}_\text{LANet}{+}\textbf{x}_\text{SENet}
\end{equation}

The combined module consisting of a LANet, a SENet, and a residual connection as three parallel branches is named as a LASE-Net. In a LASE-Net module, the LANet devotes to channel attention, while SENet devotes to spatial attention, thus both channel attention and spatial attention are employed, resulting into a better feature matrix. The residual connection facilitates the training process for better convergence. 

\subsection{Dedicated Branches for Fatigue Detection and Face Recognition}

The fatigue detection branch consists of one LASE-Net module followed by one fully connected layer. The shared feature matrix outputted by the root module undergoes feature extraction through the LASE-Net, then passes the fully connected layer to obtain the fatigue detection result of yes or no.

Similarly, the face recognition branch also consists of one LASE-Net module and one fully connected layer. The shared feature matrix outputted by the root module undergoes feature extraction through the LASE-Net, then is mapped to a 512-dimensional feature vector by the fully connected layer. The cosine distance is calculated between this feature vector and the feature vectors in the face database to determine its identity.

\section{Training Techniques for T-SCAF}
\label{sec_train}
Existing datasets are initially constructed and labeled for one dedicated task. To train our T-SCAF multi-task model using existing single-task datasets, we adopted two techniques including alternating updation and gradient accumulation. 

\subsection{Alternating Updation}
This technique is inspired by GAN. A GAN model contains two components: a generator and a discriminator. The generator generates a batch of fake samples, which are fed into the discriminator along with real samples. The discriminator's objective is to distinguish between real and fake samples and compute the loss to update its parameters. The generator's objective is to make the discriminator misclassify the samples it generated as real one, and based on this to compute the loss to update the generator's parameters. These two steps alternate until the model converges.

In our multi-task model for fatigue detection and face recognition, we also employ alternating parameter updation. When updating the fatigue detection branch, fatigue detection data is sampled and fed into the network for forward and backward propagation. The loss and gradients are computed based on the output of the fatigue detection branch, and the parameters of the root module and the fatigue detection branch are updated whereas the face recognition branch is frozen. When updating the face recognition branch, face recognition data is sampled and fed into the network. The loss and gradients are computed based on the output of the face recognition branch, and the parameters of the root module and the face recognition branch are updated whereas the fatigue detection branch is frozen. These two steps alternate until the model converges.

\subsection{Gradient Accumulation}
\label{3.2}
Gradient accumulation is initially devised to train models with large batch size under limited hardware resources. The main idea is to partition a large batch of data into smaller batches. During each iteration, a small batch of data is processed for forward and backward propagation to compute the gradients. However, instead of immediately updating the model parameters, the gradients are accumulated. After processing all the small batches, the accumulated gradients are then used to update the model parameters, thereby mimicking the training with a large batch size.

We apply this concept to the training of a multi-task model. Currently, we only have labeled datasets for single tasks: fatigue detection and facial recognition. In each step, we take one batch of data for fatigue detection and one batch of data for facial recognition. We then perform forward and backward propagation for each task independently, accumulating gradients without updating model parameters. After completing forward and backward propagation for both tasks, we compute the accumulated gradients and update the model parameters together. This approach enables us to train a multi-task model using single-task datasets.

We use cross-entropy loss in the fatigue detection branch, as shown in Eq.~\eqref{eqoen}, where $N$ is the total number of samples, $y_i$ is the label value, and $p_i$ is the probability predicted by the network.
\begin{equation}
\label{eqoen}
L_\text{Drowsy}{=}\frac{1}{N}\sum_{i}{-}[y_{i}{\cdot}\log(p_{i}){+}(1{-}y_{i}){\cdot}\log(1{-}p_{i})].
\end{equation}

In the face recognition branch, ArcFace loss\cite{arcface} is employed, which is defined in Eq.~\eqref{mlpen}.

\begin{subequations}
\label{mlpen}
\begin{equation}
\label{mlpen_1}
L_\text{Face}={-}\log\frac{e^{s\cdot\cos\left(\theta_{y_i}{+}m\right)}}{e^{s\cdot\cos\left(\theta_{y_i}{+}m\right)}{+}\sum_{j{=}1,j{\neq}y_i}^Ne^{s\cdot\cos\theta_j}}.
\end{equation}
\begin{equation}
\label{mlpen_2}
\theta_{j}{=}\arccos\bigl(\max_{k}\bigl(\textbf{W}_{i_{k}}^\text{T}\mathbf{x}_{i}\bigr)\bigr), k{\in}\{1,\cdots,K\}, W{\in}\mathbb{R}^{512{\times}N{\times}K}
\end{equation}
\end{subequations}

In Eq.~\eqref{mlpen}, $\theta_{y_i}$ denotes the angle between the extracted feature vector and the true label; $m$ is the penalty hyper-parameter set to 0.5; $s$ is the scaling parameter set to 20; $N$ is the number of classes for training, and $K$ is a hyper-parameter set to 3. $\textbf{W}^\text{T}$ represents the transpose of matrix $\textbf{W}$. $\textbf{W}_i$ represents the $i$-th column vector of matrix $\textbf{W}$. We set the weight of the fatigue detection loss to $w$ and the face recognition loss to $1{-}w$. Therefore, our overall optimization objective is expressed in Eq.~\eqref{lg}.
\begin{equation}
\label{lg}
    L{=}w{\cdot}L_\text{Drowsy}{+}(1{-}w){\cdot}L_\text{Face}.
\end{equation}

\section{Performance Evaluation}
\label{sec_evaluation}
In this section, we first introduce the dataset used, and then provide experiment details. To validate the effectiveness of our method, we present the results of ablation experiments. As there are no publicly available works specifically addressing the joint tasks of fatigue detection and face recognition, we compare our method with other single-task methods for fatigue detection and face recognition.

\subsection{Datasets}
We trained our model on a self-built fatigue detection dataset and the CASIA-Webface dataset~\cite{yi2014learning}. We validated the fatigue detection branch and face recognition branch respectively on our self-built fatigue detection dataset and the Labeled Faces in the Wild (LFW) dataset~\cite{Huang2008LabeledFI}. 

\textbf{Self-built Dataset:} The self-built dataset includes two types of video recordings: (1) surveillance footage from buses operating in Xiamen, China and (2) simulated driving videos recorded with RGB cameras in stationary vehicles. 
Surveillance footage contains the driving experiences of 32 drivers, including videos in a diverse range of conditions. Videos from the bus surveillance systems, taken with grayscale cameras, span daytime and nighttime driving, present a realistic spectrum of lighting conditions. 
The simulated driving videos, recorded with RGB cameras, represent daytime driving scenarios. Our dataset further includes challenging situations, such as low illumination, vital facial features such as the mouth or eyes concealed. 

We extracted images from the video data at intervals of 10 frames and labeled each frame as either 'Drowsy' or 'No Drowsy'. Subsequently, these labeled images were systematically segregated into a training set and a test set at a 9:1 ratio to facilitate the development and evaluation of our analytical model.

\begin{figure}[ht]
    \centering
    \includegraphics[width=1\linewidth]{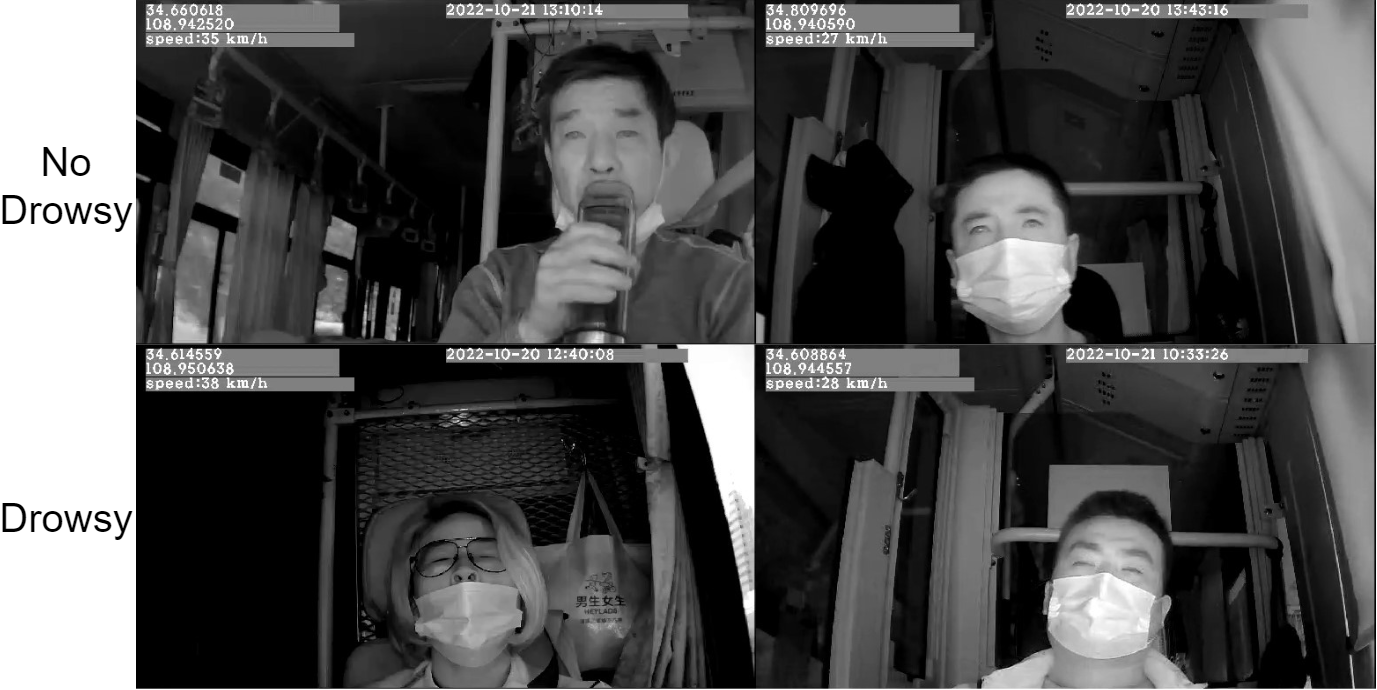}
    \caption{Images in drowsy and no-drowsy status in the self-built dataset}
    \label{fig:dataset}
\end{figure}

Fig.~\ref{fig:dataset} shows some sample pictures of drivers in our dataset in both drowsy and no-drowsy status.

\textbf{CASIA-WebFace~\cite{yi2014learning}}: CASIA-WebFace is a large-scale face recognition dataset created and maintained by the Institute of Automation, Chinese Academy of Sciences. One of its notable features is the inclusion of 494,414 face images sourced from the Internet, covering over 10,000 unique identities. This dataset exhibits diversity in facial expressions, poses, and lighting conditions, making it a challenging benchmark for testing. Each image comes with corresponding identity labels, providing information about the individuals depicted, thus making the dataset suitable for training and evaluating face recognition methods.

\textbf{Labeled Faces in the Wild~\cite{Huang2008LabeledFI}}: Labeled Faces in the Wild (LFW) is a public dataset frequently used for face recognition research. It comprises over 13,000 real-world face images collected from the Internet, each accompanied by corresponding identity labels. Widely employed for evaluating the performance of face recognition methods, LFW presents challenges such as diverse facial expressions, occlusions, and varying lighting conditions, reflecting real-world complexity. The common evaluation protocol for this dataset is one-to-one mode, which involves determining whether a pair of face images depict the same individual.

\subsection{Implementation Details}
All models were implemented using the PyTorch deep learning framework in Python. Throughout our experiments, training and testing were conducted on a computer equipped with an RTX 3090 GPU. The computer features an AMD Ryzen Threadipper Pro 3995WX CPU and 64GB of RAM, running Ubuntu 22.04. 

\begin{figure}[ht]
\centering
\includegraphics[width=1\linewidth]{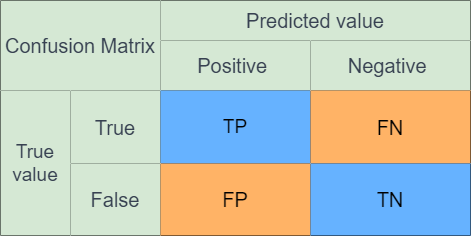}
\caption{Confusion Matrix}
\label{fig:TPFP-label}
\end{figure}

For a 2-class classification method, we use True Positive (TP) to denote the number of cases correctly predicted as positive, False Positive (FP) to indicate the number of cases wrongly predicted as positive, True Negative (TN) to represent the number of cases correctly predicted as negative, and False Negative (FN) to signifie the number of cases wrongly predicted as negative. Fig.~\ref{fig:TPFP-label} shows the relationships among them.

We utilized accuracy (ACC) to evaluate the fatigue detection performance of the T-SCAF model. ACC can be calculated as Eq.~\eqref{acc}.

\begin{equation}
\label{acc}
\text{ACC}=\frac{\text{TP}+\text{TN}}{\text{TP}+\text{TN}+\text{FP}+\text{FN}}
\end{equation}

As for experiment details, we utilized RetinaFace\cite{retinaface} to extract facial regions from the original images and manually adjusted them for calibration. All extracted facial images were resized to $112{\times}112$ pixels. For the root module of our T-SCAF model, we used IResNet50\cite{iresnet}, which is the backbone network of ArcFace\cite{arcface}. We utilized the Adaptive Moment Estimation optimizer (ADAM) to optimize the parameters, with an initial learning rate of 1e-2 and momentum of 0.9. The model was trained for 20 epochs with a batch size of 512, and the learning rate was adjusted every 20 epochs using CosineAnnealingLR. For both gradient accumulation and alternating updation, we set the data sampling method to sample a batch of data alternately from the fatigue detection dataset and the face recognition dataset.

\subsection{Comparative Experiment}

We compared our model with single-task models for fatigue detection and face recognition, and the quantitative results are shown in Table~\ref{tab:4} and Table~\ref{tab:4_2}. 

Compared to the methods in \cite{real-time,32dua2021deep,31park2016driver,33husain2022development}, our model achieved an accuracy improvement of 3.2\%, 0.61\%, 0.96\%, and 1.18\% respectively in the fatigue detection task. Compared to ArcFace, our model's accuracy decreased by 0.77\% in the face recognition task. Compared to combination methods combining each of \cite{real-time,31park2016driver,33husain2022development} and the ArcFace, our model achieved an average accuracy improvement of 1.18\%, 0.1\%, 0.21\% respectively in both tasks. It is noteworthy that our model achieved the results with fewer parameters and computations than the three aforementioned single-task model combination methods. 

Compared to a method combining the one in~\cite{32dua2021deep} and ArcFace, our model's average accuracy decreased by 0.08\%, but our model has only 1/7 parameters and requires only 1/5 computations. Compared to the combination method of our previously proposed fatigue detection model MAF~\cite{qu2023multi} and ArcFace, our T-SCAF model's average accuracy decreased by 0.77\%, but nearly half parameters and computations are saved. Additionally, we replaced the root backbone module of T-SCAF model with the more lightweight MobileNetV2~\cite{37mobilenet}, resulting into a mode with 1/3 parameters and 1/50 computations of the previous model using IResNet50\cite{iresnet}, whereas with a trivial penalty of accuracy reduction of 4.77\% . 

\begin{table*}[]
\centering
\begin{tabular}{c|cc|cc}
\hline
& \textbf{\makecell[c]{Fatigue Detection Model}}
& \textbf{\makecell[c]{Face Recognition Model}} 
& \textbf{\makecell[c]{Model Size(M)$\downarrow$\\}}
& \textbf{\makecell[c]{GFLOPs$\downarrow$}}\\ 
\hline
\multirow{5}{*}{\makecell[c]{Parallel-style \\multi-task model}} 
& Method\cite{real-time} & ArcFace\cite{arcface}     & 53.48         & 17.00 \\
& Method\cite{32dua2021deep}  & ArcFace\cite{arcface}         & 315.39  & 64.73 \\
& Method\cite{31park2016driver}   & ArcFace\cite{arcface}    & 157.61   & 13.89  \\
& Method\cite{33husain2022development} & ArcFace\cite{arcface} & 85.72  & 16.81 \\
& MAF\cite{qu2023multi}  & ArcFace\cite{arcface}             & 110.09   & 25.49 \\ \hline
\multirow{2}{*}{\makecell{Tree-style \\multi-task model}}  
& \multicolumn{2}{c|}{Ours(MobileNetV2)}  & \textbf{\color{red}{14.77}}   & \textbf{\color{red}{0.25}}  \\
& \multicolumn{2}{c|}{Ours(IResNet50)}    & \color{blue}{43.92}   & \color{blue}{12.72} \\ 
\hline
\end{tabular}
\caption{Model size and FLOPs of different models}
\label{tab:4}
\end{table*}

\begin{table*}[]
\centering
\begin{tabular}{c|cc|ccc}
\hline
& \textbf{\makecell{Fatigue Detection\\Model}}
& \textbf{\makecell{Face Recognition\\Model}}  
& \textbf{\makecell{ACC$\uparrow$\\(Fatigue Detection)}}
& \textbf{\makecell{ACC$\uparrow$\\(Face Recognition)}} 
& \textbf{\makecell{ACC$\uparrow$\\(Avg)}} \\ \hline
\multirow{5}{*}{\makecell{Parallel-style\\ multi-task model}} 
& Method\cite{real-time}     & ArcFace\cite{arcface} & 0.9297  & 0.9620    & 0.9459  \\
& Method\cite{32dua2021deep} & ArcFace\cite{arcface} & 0.9549 & 0.9620  & \color{blue}{0.9585}  \\
& Method\cite{31park2016driver} & ArcFace\cite{arcface}  & 0.9514  & 0.9620 & 0.9567 \\
& Method\cite{33husain2022development}  & ArcFace\cite{arcface}  & 0.9492  & 0.9620   & 0.9556  \\
& MAF\cite{qu2023multi} & ArcFace\cite{arcface}   & \textbf{\color{red}{0.9688}} & \textbf{\color{red}{0.9620}}  & \textbf{\color{red}{0.9654}} \\ 
\hline
\multirow{2}{*}{\begin{tabular}[c]{@{}c@{}}Tree-style \\ multi-task model\end{tabular}}                 & \multicolumn{2}{c|}{Ours(MobileNetV2)}  & 0.9091  & 0.9108  & 0.9100\\
& \multicolumn{2}{c|}{Ours(IResNet50)}   & \color{blue}{0.9610}  & \color{blue}{0.9543}  & 0.9577\\ 
\hline
\end{tabular}
\caption{ACC of various models}
\label{tab:4_2}
\end{table*}

\subsection{Ablation Experiments}

\begin{table*}[ht]
\centering

\begin{tabular}{ccc|ccc}
\hline
\textbf{Backbone} & \textbf{SENet} & \textbf{LANet} 
& \textbf{ACC(Fatigue Detection)$\uparrow$} 
& \textbf{ACC(Face Recognition)$\uparrow$} 
& \textbf{ACC(Avg)$\uparrow$} \\
\hline
$\checkmark$        & $\checkmark$     &       & 0.9318   & 0.9033  & 0.9176   \\
$\checkmark$        &       & $\checkmark$     & \color{blue}{0.9578}   & \color{blue}{0.9465}  & \color{blue}{0.9522}   \\
$\checkmark$        & $\checkmark$     & $\checkmark$     
& \textbf{\color{red}{0.9610}} & \textbf{\color{red}{0.9543}} & \textbf{\color{red}{0.9577}}\\ 
\hline
\end{tabular}
\caption{Results of ablation experiments}
\label{tab:ad}
\end{table*}

We conducted ablation experiments by progressively removing modules from the model to validate the effectiveness of each module. The quantitative results of the ablation experiments are shown in Table \ref{tab:ad}. We added attention modules to both branches to separate the features that are more useful for each task extracted from the root backbone module. We found that When only adding the LANet module, the accuracy of fatigue detection and face recognition tasks increased by 1.95\% and 0.25\%, respectively. When adding both SENet and LANet simultaneously, the inclusion of SENet lead to futher performance improvement, with the accuracy of the fatigue detection and face recognition tasks increased by 2.27\% and 1.03\%, respectively. However, when only adding the SENet module, both the accuracy of fatigue detection and face recognition tasks decreased with varying degrees, by 0.65\% and 4.13\%, respectively. This may because that, channel attention ability of SENet reduces the feature weights across the entire channel dimension, which may impair the effectiveness of the features. 

We also compared the two training methods of alternating updation and gradient accumulation. The quantitative results are shown in Table~\ref{tab:2}. The alternating updation method achieved the highest accuracy when the two tasks are combined. The gradient accumulation training method allows controlling the influences of the two tasks on model updating by adjusting the loss weight. With an increase in $w$, the overall accuracy of fatigue detection tends to improve, while the accuracy of face recognition first increases and then decreases. The comprehensive accuracy of fatigue detection and face recognition reaches its maximum when $w$ is set to 0.5. Additionally, we found that gradient accumulation requires larger GPU memory resources compared to alternating updation, as it involves storing the results of two backward passes during training.

\begin{table*}[]
\centering
\begin{tabular}{cc|ccc}
\hline
\multicolumn{2}{c|}{\textbf{Training techniques}} 
& \textbf{ACC(Fatigue Detection)$\uparrow$} 
& \textbf{ACC(Face Recognition)$\uparrow$} 
& \textbf{ACC(Avg)$\uparrow$}        \\ 
\hline
\multicolumn{2}{c|}{Alternating updation} & \textbf{\color{red}{0.9610}}  & \textbf{\color{red}{0.9543}} & \textbf{\color{red}{0.9577}} \\ 
\hline
\multicolumn{1}{c|}{\multirow{10}{*}{Gradient accumulation}} & $w$ & \multicolumn{3}{c}{} \\ 
\cline{2-5} \multicolumn{1}{c|}{} & 0.1& 0.8994 & 0.9318 & 0.9156 \\
\multicolumn{1}{c|}{}  & 0.2 & 0.8506  & 0.9085 & 0.8796 \\
\multicolumn{1}{c|}{}  & 0.3 & 0.8474  & 0.9107 & 0.8791 \\
\multicolumn{1}{c|}{}  & 0.4 & 0.9286  & 0.9412 & 0.9359 \\
\multicolumn{1}{c|}{}  & 0.5 & 0.9545  & \color{blue}{0.9473} & \color{blue}{0.9509} \\
\multicolumn{1}{c|}{}  & 0.6 & 0.9513  & 0.9243 & 0.9378 \\
\multicolumn{1}{c|}{}  & 0.7 & 0.9513  & 0.9117 & 0.9315 \\
\multicolumn{1}{c|}{}  & 0.8 & 0.9513  & 0.9270 & 0.9392 \\
\multicolumn{1}{c|}{}  & 0.9 & \textbf{\color{red}{0.9610}}  & 0.9183 & 0.9397 \\ 
\hline
\end{tabular}
\caption{Impacts of loss weights of alternating updation and gradient accumulation training techniques.}
\label{tab:2}
\end{table*}

Furthermore, to verify that the root backbone module extracts shared universal features that can be used for both fatigue detection and face recognition tasks, we split the two tasks in the multi-task model into two separate modules, each composed of the same copy of the backbone module and its corresponding LASE-Net branch module. The quantitative results are shown in Table~\ref{tab:3}. Compared with parallel-style multi task model, our T-SCAF multi-task model achieves improvements in accuracy respectively by 0.4\% and 0.33\% in the fatigue detection and face recognition tasks. The results indicates that, the T-SCAF multi-task model enables both tasks to share underlying feature representations, allowing the model to learn universal features. Additionally, each task in the multi-task model can be considered as a regularization term for the other task. As the model needs to find a balance between multiple tasks, it tends to mitigate over-fitting and improve generalization performance on unseen data, thus achieving better results on the test dataset.

\begin{table*}[]
\centering
\begin{tabular}{ccc}
\hline
\textbf{Model}                                
& \textbf{ACC(Fatigue Detection)$\uparrow$} 
& \textbf{ACC(Face Recognition)$\uparrow$} \\ 
\hline
Backbone(root module) + Fatigue detection branch  & 0.9570   &        \\
Backbone(root module) + Face recognition branch &          & 0.9510  \\ 
\hline
Ours  & \textbf{\color{red}{0.9610}} & \textbf{\color{red}{0.9543}} \\ 
\hline
\end{tabular}
\caption{Results of splitting our T-SCAF multi-task model into parallel-style multi-task model.}
\label{tab:3}
\end{table*}

\subsection{Result Visualization}

To further validate the effectiveness of the T-SCAF multi-task model, we visualized the features of some intermediate layers.

\begin{figure}
    \centering
    \includegraphics[width=1\linewidth]{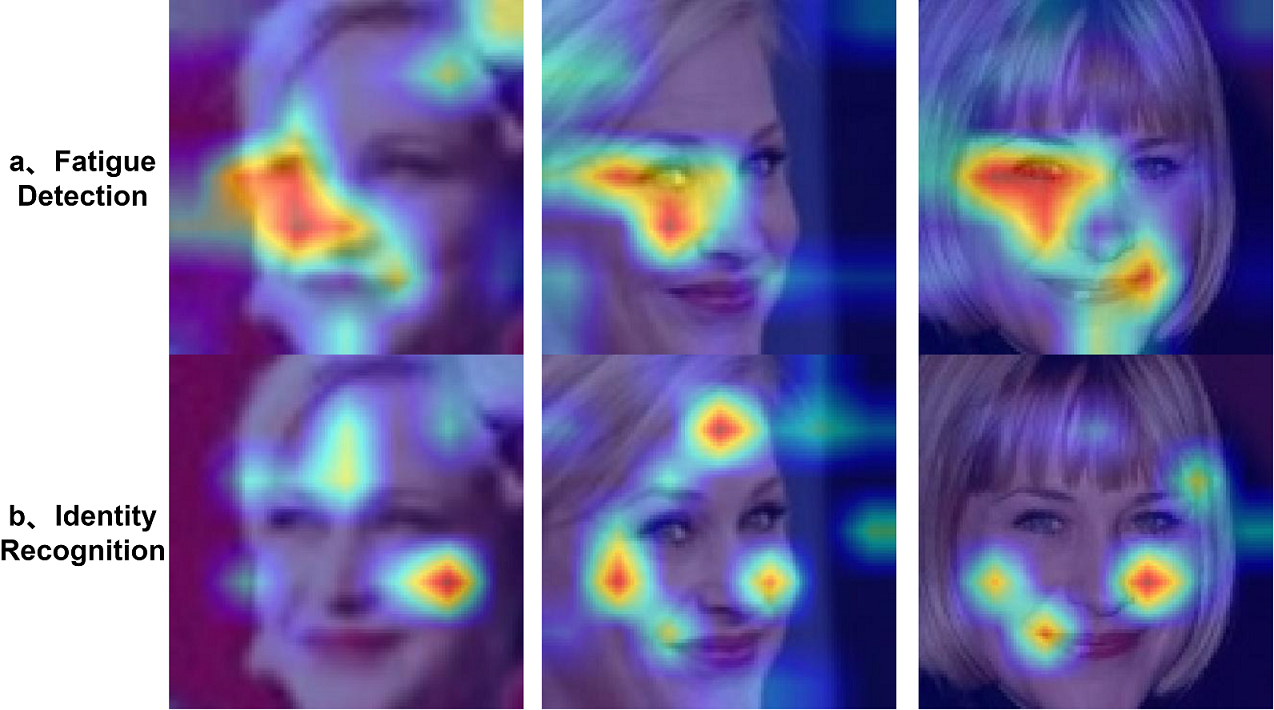}
    \caption{Visualization of LASE-Net output feature maps}
    \label{fig:cam}
\end{figure}

The feature matrices returned by the LASE-Net module in the fatigue detection branch and face recognition branch using the Grad-CAM method\cite{42cam} are shown in Fig.~\ref{fig:cam}. The first row depicts the visualization of the feature matrix returned by the LASE-Net module in the fatigue detection branch, whereas the second row shows those of the face recognition branch. It can be observed that, for different detection tasks, the two branches focus on different facial regions from the shared features, which are extracted by the root backbone module. The features emphasized by the fatigue detection branch are concentrated around the eyes and mouth regions, whereas the features emphasized by the face recognition branch are dispersed across different parts of the face.

\begin{figure}
    \centering
    \includegraphics[width=1\linewidth]{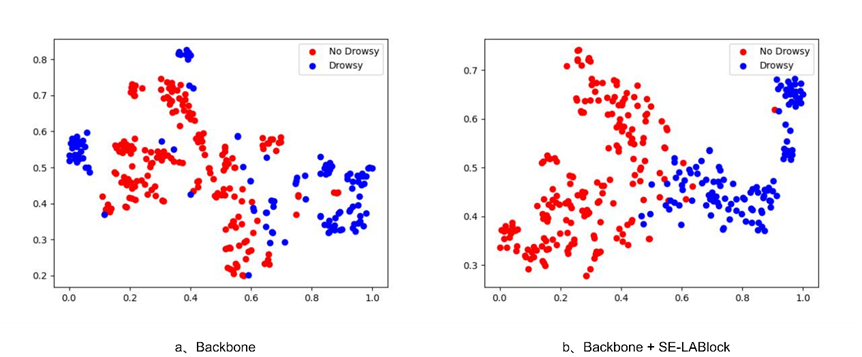}
    \caption{t-SNE visualization results.}
    \label{fig:sne}
\end{figure}

We visualized the feature vectors before the last fully connected layer in the fatigue detection branch of two model configurations, one with only the root backbone module, the other combines the root backbone module and an LASE-Net module. These configurations are respectively referred as Backbone multi-task model and Backbone+LASE-Net multi-task model. We used the t-SNE\cite{t_sne} technique to embed high-dimensional features in a two-dimensional plane for better interpretation. The visualization results are shown in Fig.~\ref{fig:sne}, with the left image representing the Backbone multi-task model and the right image representing the Backbone+LASE-Net multi-task model. The results show that, compared to the Backbone multi-task model, the Backbone+LASE-Net multi-task model exhibits clearer boundaries between different categories and tighter aggregation within the same category. This indicates that the LASE-Net module enhances the feature extraction capability of the fatigue detection branch.

\begin{figure}
    \centering
    \includegraphics[width=1\linewidth]{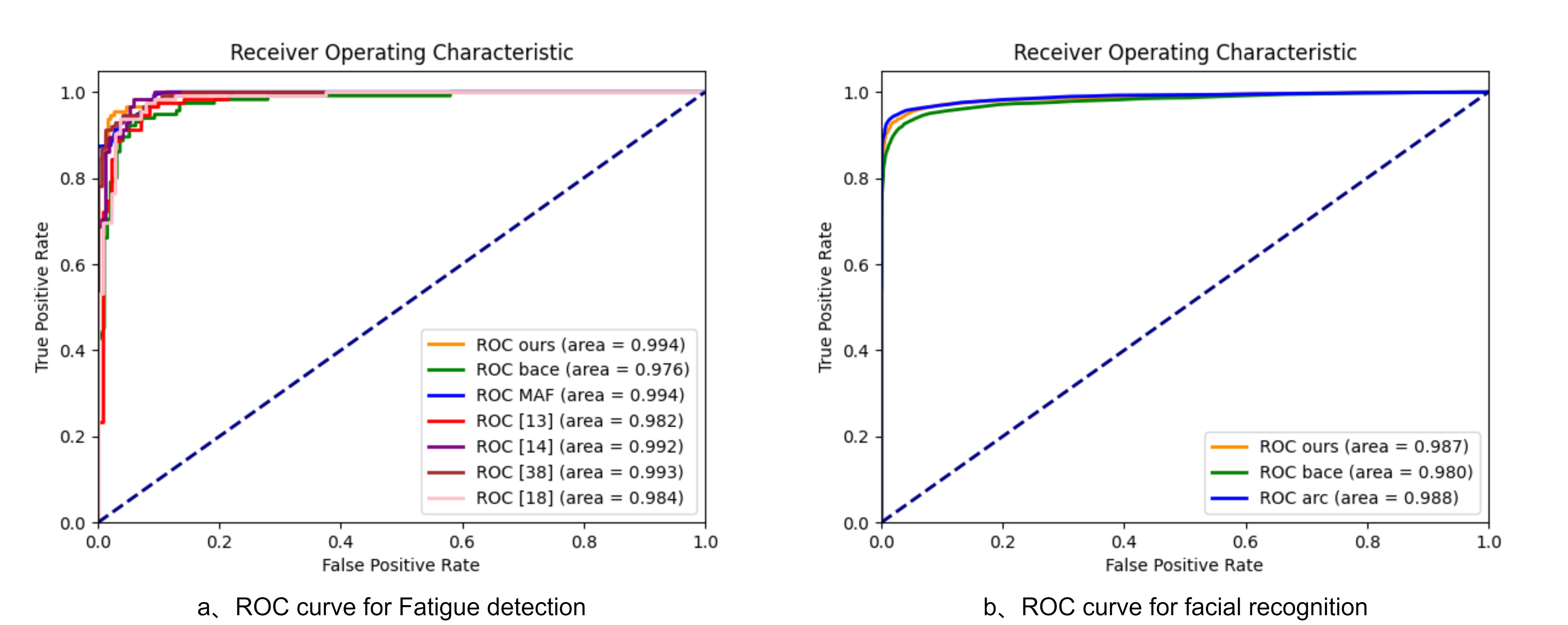}
    \caption{ROC curve comparison of different models}
    \label{fig:roc}
\end{figure}

The ROC curves for fatigue detection and face recognition tasks of our T-SCAF multi-task model and the other comparison models are shown in Fig. \ref{fig:roc}. The Area Under Curve (AUC) values for each model in different tasks are annotated along the curve. In the fatigue detection task, T-SCAF's AUC value is 0.022 higher than that of the Basic model (which only includes the shared root backbone module without the LASE-Net). It shares the same AUC value with the single-task model MAF\cite{qu2023multi} and shows varying degrees of improvement compared to other single-task models. In the face recognition task, T-SCAF's AUC value is 0.004 higher than that of the Basic model. Additionally, even with a lighter model size and faster computation speed, T-SCAF's AUC is only 0.001 lower than the single-task models. This demonstrates the effectiveness of our T-SCAF multi-task model.

\section{Conclusion}
\label{sec_conclusion}
This paper proposes a novel tree-style multi-task modeling approach for multi-task learning. which rooted at a shared backbone, more dedicated separate module branches are appended as the model pipeline goes deeper. Following the tree-style approach, we propose a tree-style space-channel attention fusion (T-SCAF) network model for simultaneously performing driver fatigue detection and face recognition for identifying a driver. This model shares a common feature extraction backbone module, with further separated feature extraction and classification module branches. The dedicated branches exploit and combine spatial and channel attention mechanisms to generate space-channel fused-attention enhanced features, leading to improved detection performance. we also introduce alternating updation and gradient accumulation techniques for training our T-SCAF model using only the single-task datasets. The experiments on our self-built dataset, CASIA-WebFace, and LFW dataset validates the effectiveness of our T-SCAF model. 



\printcredits

\bibliographystyle{ieeetr}
\bibliography{article}

\end{sloppypar}
\end{document}